\documentclass{article}
\usepackage{leaplab}
\usepackage[table,x11names]{xcolor}

\usepackage[utf8]{inputenc} 
\usepackage[T1]{fontenc}    
\usepackage{url}            
\usepackage{booktabs}       
\usepackage{amsfonts}       
\usepackage{nicefrac}       
\usepackage{microtype}      
\usepackage{tabularx}   
\usepackage{enumitem}

\usepackage[utf8]{inputenc} 
\usepackage[T1]{fontenc}    
\usepackage{url}            
\usepackage{booktabs}       
\usepackage{amsfonts}       
\usepackage{nicefrac}       
\usepackage{microtype}      
\definecolor{citecolor}{HTML}{0071bc}
\definecolor{shadecolor}{HTML}{EFEFEF}
\definecolor{linkcolor}{HTML}{9A4D92}
\usepackage[colorlinks, linkcolor=linkcolor, colorlinks, anchorcolor=blue, citecolor=citecolor, pagebackref=True]{hyperref}

\usepackage{amsmath}
\usepackage{microtype}
\usepackage{graphicx}
\usepackage{booktabs} 
\usepackage{caption}
\usepackage{wrapfig}
\usepackage{float}

\usepackage{titletoc} 

\usepackage{subcaption}
\usepackage[export]{adjustbox}
\usepackage{footnote}
\usepackage{tablefootnote}
\usepackage{threeparttable}
\usepackage[capitalize,noabbrev]{cleveref}

\usepackage{tabularray}
\UseTblrLibrary{booktabs}
\usepackage{multirow}
\usepackage{multicol}
\usepackage{xspace}
\usepackage{mathrsfs}
\usepackage{pifont}
\usepackage{makecell}
\usepackage{CJK}

\usepackage{marvosym} 
\usepackage[most]{tcolorbox}

\definecolor{brand}{HTML}{000000}
\definecolor{xulabwine}{HTML}{5E3545}
\definecolor{titlebg}{HTML}{F7F2F3}

\hypersetup{citecolor=citecolor,linkcolor=linkcolor,filecolor=linkcolor,urlcolor=magenta}

\usepackage{titletoc}

\newenvironment{leapabstract}{\begin{tcolorbox}[
colback=titlebg,colframe=white,boxrule=0pt,arc=10pt,
left=16pt,right=16pt,top=12pt,bottom=12pt,width=\textwidth,
enlarge left by=0mm,before skip=10pt,after skip=10pt]}{\end{tcolorbox}}

\fancypagestyle{fancytitlepage}{\fancyhead{}\lhead{\color{brand}\Large\bfseries Xu Lab}\rhead{\color{brand}\itshape HKUST}\cfoot{}}

\usepackage{amssymb}
\usepackage{textcomp}
\usepackage{placeins}
\usepackage{needspace}
\newcommand{\cmark}{\textcolor{green!50!black}{\ding{51}}}
\newcommand{\xmark}{\textcolor{red!80!black}{\ding{55}}}
\icmltitlerunning{FOMI: A Force-Feedback Manipulation Interface for Teaching Contact-Rich Manipulation}

\begin{document}
\icmltitle{FOMI: A Force-Feedback Manipulation Interface for Teaching Contact-Rich Manipulation}
\begin{icmlauthorlist}
Yishu~Shen$^{1}$,
Yiming~Luo$^{2\dagger}$,
Ruilin~Wang$^{3}$,
Bowen~Yang$^{3}$,
Shuai~Yang$^{1}$,
Fangyi~Xu$^{3}$,
Yujun~Shen$^{3}$,
Fu~Zhang$^{2}$,
Boyu~Zhou$^{4}$,
and Yinghao~Xu$^{1*}$
\end{icmlauthorlist}
\par\addvspace{6pt}

{\small
\setlength{\parskip}{0pt}
$^{1}$The Hong Kong University of Science and Technology\par
$^{2}$The University of Hong Kong\par
$^{3}$Robbyant\par
$^{4}$Southern University of Science and Technology\par
$^{\dagger}$Project leader.\quad
$^{*}$Corresponding author
(e-mail: \texttt{justimyhxu@ust.hk}).\par
}
\vskip .15in

\begin{leapabstract}
Robot-free collection with handheld interfaces enables scalable acquisition of human demonstrations for policy learning.
However, existing handheld interfaces often provide limited intuitive force feedback, making contact-rich demonstration collection challenging for human operators.
To address this, we present FOMI, a force-feedback manipulation interface with a scissor-shaped mechanism, which transmits object interaction forces directly to the operator's hands through actuation links without force sensors. 
This intuitive force feedback supports real-time action adjustment during robot-free collection of contact-rich demonstrations. 
Around FOMI, we build a demonstration collection system that integrates sensor calibration, data acquisition, and synchronization for robot-free collection.
By further mapping pose changes and gripper opening of FOMI to robot end-effector poses and gripper commands, the system also supports teleoperation.
Across four contact-rich tasks, FOMI improves collection efficiency by approximately 29\% and 18\% over the best baselines in robot-free collection and teleoperation, respectively.
Demonstrations collected with FOMI in both modes enable
policies to perform challenging contact-rich manipulation tasks with
high success rates.


\par\addvspace{6pt}
{\centering\small
\textbf{Project page:}~\url{https://robbyant-research.github.io/FOMI/}\par
}
\end{leapabstract}

\newcommand{\FOMIteaser}{%
  \begin{figure}[!htbp]
    \centering
    \includegraphics[width=0.99\textwidth]{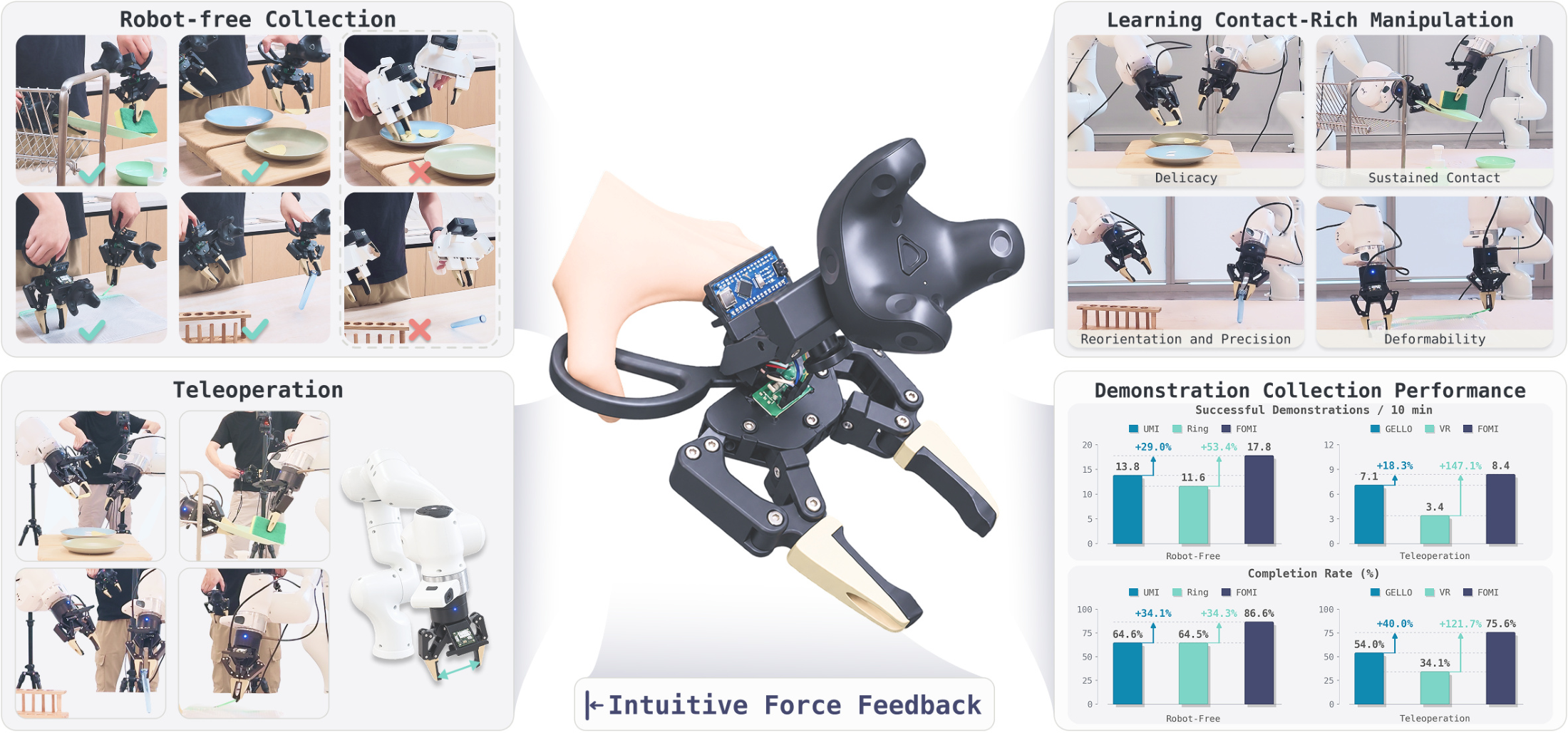}
    \par
    \caption{Overview of FOMI. The force-feedback
    manipulation interface incorporates a scissor-shaped mechanism
    to provide human operators with intuitive force feedback in
    robot-free collection of contact-rich demonstrations, while
    also supporting teleoperation for policy learning. FOMI achieves
    higher collection efficiency and completion rates than the
    evaluated baselines in both modes. Policies fine-tuned on
    demonstrations collected with FOMI can perform contact-rich
    manipulation tasks in the real world.}
    \label{fig:teaser}
  \end{figure}
}

\FOMIteaser

\section{Introduction}

Human demonstrations provide behavioral supervision for manipulation policy learning. 
Teleoperation~\cite{wu2024gello,zhao2023aloha} collects demonstrations directly on the target robot but requires robot hardware, limiting data scale and diversity.
Robot-free collection with handheld interfaces such as the Universal Manipulation Interface (UMI)~\cite{chi2024umi,fastumi2024} has therefore been widely explored for scalable demonstration acquisition in diverse real-world environments.
However, collecting contact-rich manipulation demonstrations with these handheld interfaces remains challenging because a lack of intuitive force feedback can limit human operators' ability to perceive object interaction forces and adjust actions in real time~\cite{liu2025forcemimic}, resulting in more failed attempts and lower collection efficiency~\cite{wu2025freetacman}.

A key factor is the interface's mechanical design, which determines how directly interaction forces are transmitted to the human operator.
Common handheld interfaces in robot-free demonstration collection employ trigger-driven rack-and-pinion mechanisms~\cite{chi2024umi,choi2026umift}.
These mechanisms transmit object interaction forces through multiple intermediate components, where mechanical backlash weakens force feedback~\cite{wu2025freetacman}.
Alternatively, interfaces with active force feedback~\cite{zhang2025doglove,citron2026modpack} combine force sensing with motorized actuation to convey interaction forces to operators, requiring additional sensing hardware, actuators, and feedback control.
These limitations motivate a mechanical interface that provides intuitive force feedback for real-time action adjustment during contact-rich demonstration collection without additional force sensors or active feedback actuators.

Inspired by the comfortable handling and intuitive force feedback of scissors, we design a scissor-shaped mechanism to support contact-rich demonstration collection.
Each handle forms a rigid assembly with its sector gear and actuation link, preserving a direct mechanical path for force transmission without additional transmission stages between the handle and the actuation link.
To match the target parallel gripper, the two identical sector gears mesh with each other, constraining the two branches to open and close symmetrically.


Building on this mechanical design, we introduce FOMI, a force-feedback manipulation interface, together with a demonstration collection system that integrates sensor calibration, data acquisition, and synchronization for robot-free collection of contact-rich demonstrations.
Through calibration and mapping, the system further converts sensor measurements of FOMI's pose changes and gripper opening into robot end-effector pose targets and gripper width commands, enabling teleoperated demonstration collection with the same interface.
Our main contributions are as follows:

\begin{itemize}
    \item \textbf{Mechanical design for intuitive force feedback:}
    We design a scissor-shaped interface in which each handle is rigidly integrated with its actuation link, providing a direct force path for intuitive force feedback during robot-free collection, with meshing sector gears matching the symmetric motion of parallel grippers.

    \item \textbf{Demonstration collection system for teaching contact-rich manipulation:}
    We build a collection system around FOMI that integrates sensor calibration, data acquisition, and synchronization for robot-free collection, and further supports teleoperation by mapping FOMI's motion and gripper opening to robot commands.

    \item \textbf{Improved collection efficiency and policy evaluation:}
    Compared with baseline interfaces in both collection modes, FOMI achieves the highest collection efficiency and improves completion rates.
    Policies fine-tuned on FOMI demonstrations achieve high success rates on four contact-rich tasks.
    Hardware designs, code, and datasets will be released.
\end{itemize}


\section{Related Work}

\subsection{Demonstration Collection Interfaces}

Human demonstrations for policy learning are commonly collected through robot-free collection or teleoperation. In robot-free collection, UMI~\cite{chi2024umi} and its variants~\cite{choi2026umift,fastumi2024} enable demonstration collection across diverse real-world environments. In teleoperation, leader-follower interfaces such as GELLO~\cite{wu2024gello}, ALOHA~\cite{zhao2023aloha}, and Mobile ALOHA~\cite{fu2024mobile} use joint-space mapping for arm control to collect demonstrations, whereas VR systems~\cite{george2025openvr} and smartphone systems~\cite{wang2023mimicplay,agarwal2026cobalt} provide end-effector pose control. Recent interfaces have been developed to support both robot-free collection and teleoperation~\cite{seo2025legato,surendran2026bridge}.

\subsection{Force Feedback for Demonstration Collection}

\begin{table}[!htbp]
\centering
\begin{minipage}{0.72\textwidth}
\captionsetup{font=small,skip=5pt}
\caption{Mechanical design and features of robot-free demonstration
interfaces. \protect\cmark: present; \protect\xmark: absent.}
\label{tab:interface-comparison}
\centering
\footnotesize
\setlength{\tabcolsep}{3.5pt}
\renewcommand{\arraystretch}{1.25}
\renewcommand{\tabularxcolumn}[1]{m{#1}}

\begin{tabularx}{\linewidth}{
    @{}
    >{\centering\setlength{\parskip}{1pt}\arraybackslash}m{124pt}
    >{\hsize=1.60\hsize\linewidth=\hsize
      \centering\arraybackslash}X
    >{\hsize=0.66\hsize\linewidth=\hsize
      \centering\arraybackslash}X
    >{\hsize=0.74\hsize\linewidth=\hsize
      \centering\arraybackslash}X
    @{}
}
\toprule
\textbf{Interface}
& \textbf{Mechanical Design}
& \textbf{Intuitive Force Feedback}
& \textbf{Handles for Load Support} \\
\midrule

\mbox{UMI \cite{chi2024umi},}\par
\mbox{UMI-FT \cite{choi2026umift}}
& Trigger-driven rack-and-pinion
& \xmark
& \cmark \\
\addlinespace[7pt]

\mbox{FastUMI \cite{fastumi2024},}\par
\mbox{exUMI \cite{xu2025exumi}}
& \mbox{Trigger-driven actuation}
& \xmark
& \cmark \\
\addlinespace[7pt]

\mbox{FreeTacMan \cite{wu2025freetacman}}
& \mbox{Finger-driven actuation}
& \cmark
& \xmark \\

\midrule

\textbf{FOMI (Ours)}
& \textbf{Scissor-shaped actuation linkage}
& \cmark
& \cmark \\

\bottomrule
\end{tabularx}
\end{minipage}
\end{table}

The interface's mechanical design can influence force feedback and action adjustment during demonstration collection~\cite{georgadarellis2026gripper}. Existing manipulation interfaces provide force feedback through mechanical transmission or active feedback. UMI~\cite{chi2024umi} and UMI-FT~\cite{choi2026umift} use trigger-driven rack-and-pinion mechanisms, while other handheld interfaces use trigger-driven linkages~\cite{fang2026robopocket,liu2026rdt2,xu2025exumi}. These mechanisms transmit interaction forces to the hand, but multiple transmission stages and backlash may obscure force feedback. In contrast, finger-driven interfaces~\cite{ohkawa2026yubi,wu2025freetacman} transmit interaction forces to the actuating fingers, but supporting the interface with the actuating fingers may affect force perception and action adjustment. Alternatively, interfaces with active force feedback provide bilateral gripper force feedback~\cite{luo2026omniumi} or reflect interaction forces from the follower robot~\cite{liu2025factr,zhang2025doglove,citron2026modpack}, requiring additional actuators and feedback control. FOMI integrates each handle with an actuation link to provide direct mechanical force feedback and handle support during robot-free collection. Table~\ref{tab:interface-comparison} compares the mechanical designs, force feedback, and support handles of representative interfaces.

\subsection{Contact Sensing for Policy Learning}

Recent manipulation interfaces incorporate additional tactile or force sensors to collect contact information for learning contact-rich manipulation. Some interfaces collect synchronized visual and tactile observations and use tactile representation learning to support visuo-tactile policies~\cite{xu2025exumi,liu2025vitamin,zhu2026touch}. Others record interaction forces to support force-aware policy learning and contact control~\cite{jang2026wt,choi2026umift}. FOMI provides mechanical force feedback to human operators during robot-free demonstration collection, while the collected demonstrations support policy learning for contact-rich manipulation without tactile or force inputs to the policy.


\section{Methodology}

We describe FOMI's mechanical design, kinematics, and force transmission (Sec.~\ref{sec:mechanical-design}), present the collection system for robot-free collection and teleoperation (Sec.~\ref{sec:collection-system}), and outline policy learning and deployment (Sec.~\ref{sec:policy-learning}).

\subsection{Mechanical Design}
\label{sec:mechanical-design}

\begin{figure*}[!t]
    \centering
    \includegraphics[width=0.99\textwidth]{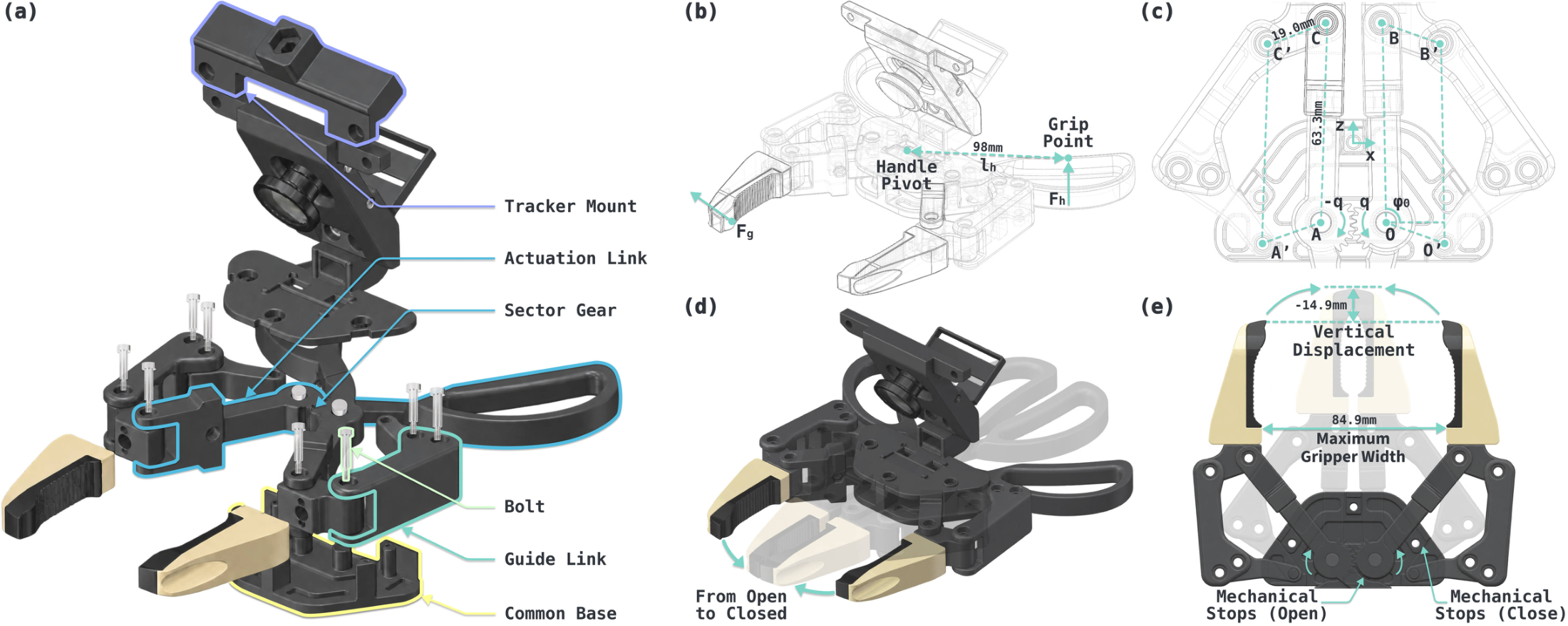}
    \caption{Mechanical design of FOMI.
    (a)~Exploded view with the main components.
    (b)~Direct force transmission path from the gripper to the handles in robot-free collection, with the handle pivot, hand force $F_h$, and effective moment arm $\ell_h$ indicated.
    (c)~Kinematic schematic showing base pivots $O,O'$ ($A,A'$) and gripper carrier joints $B,B'$ ($C,C'$) on the right (left) branch. The base coordinate axes are $x$ along the gripper opening direction, $z$ vertical, and $y$ into the page. The right and left sector gears rotate by $q$ and $-q$ from closure, respectively.
    (d)~Motion from the open to the closed configuration.
    (e)~Mechanical stops defining the gripper width limits, with the maximum gripper width $w_I(q_{\max})$ and vertical displacement $\delta(q_{\max})$ indicated.}
    \label{fig:mechanism}
\end{figure*}

\subsubsection{Scissor Mechanism}
\label{sec:mechanism}

As shown in Fig.~\ref{fig:mechanism}(a), FOMI incorporates a scissor-shaped mechanism comprising two mirrored branches supported by a common base. Each branch includes a handle, a sector gear, and an actuation link rigidly integrated into a single component. The actuation link and a guide link form a parallelogram with the gripper carrier and the base. Both links connect to the carrier through bolted revolute joints, while the parallelogram geometry keeps the gripper contact surfaces parallel to each other. The two identical sector gears have a module of $1\,\mathrm{mm}$ and a pitch radius of $10\,\mathrm{mm}$. Each gear rotates about its corresponding handle pivot, with the two pivot axes spaced $20\,\mathrm{mm}$ apart. The $1{:}1$ transmission ratio constrains the two branches to rotate through equal angles in opposite directions, producing symmetric gripper opening and closing.

In robot-free collection, interaction forces at the gripper are transmitted through the actuation links directly to the handles, providing intuitive force feedback to operators, as shown in Fig.~\ref{fig:mechanism}(b). The mechanism also has no return springs, so operators actively drive both opening and closing, and the forces they perceive arise from object interactions rather than spring preload. The handles also serve as the grip for holding and supporting the interface during manipulation.

\subsubsection{Kinematic and Force Transmission Analysis}
\label{sec:kinematic-Transmission}

As shown in Fig.~\ref{fig:mechanism}(c), we define the opening angle $q\in[0,q_{\max}]$ as the rotation of the right sector gear about $O$ from the closed position. The two parallelogram linkages satisfy
\begin{equation}
    \overrightarrow{OB}=\overrightarrow{O'B'},\qquad
    \overrightarrow{AC}=\overrightarrow{A'C'},
    \label{eq:parallelogram-constraint}
\end{equation}
which yields $\overrightarrow{BB'}=\overrightarrow{OO'}$ and $\overrightarrow{CC'}=\overrightarrow{AA'}$. Therefore, both gripper carriers maintain a constant angle relative to the base.

To match the visual geometry and opening--closing motion of the target Robotiq 2F-85 parallel gripper, we select an actuation link length of $\rho=63.3\,\mathrm{mm}$ and an initial angle of $\varphi_0=91.5^\circ$. Here, $\varphi_0$ is the angle between the actuation link $OB$ at closure and the positive $x$-axis of the base, as shown in Fig.~\ref{fig:mechanism}(c). 
During interface opening and closing shown in Fig.~\ref{fig:mechanism}(d), this angle is $\varphi_0-q$. Mechanical stops limit the motion between the fully closed and fully open positions, defining the operating range $q\in[0,41.6^\circ]$, as shown in Fig.~\ref{fig:mechanism}(e). Taking the closed position as the zero reference, the interface gripper width $w_I$ and the vertical displacement $\delta$ of two gripper midpoints are
\begin{align}
    w_I(q)&=2\rho\left[\cos(\varphi_0-q)-\cos\varphi_0\right],
    \label{eq:width-kinematics}\\
    \delta(q)&=\rho\left[\sin(\varphi_0-q)-\sin\varphi_0\right].
    \label{eq:gripper-vertical}
\end{align}
With $q$ in radians, the derivative
of gripper width with respect to $q$ is
\begin{equation}
    \frac{\mathrm{d}w_I}{\mathrm{d}q}
    =2\rho\sin(\varphi_0-q)>0.
    \label{eq:opening-jacobian}
\end{equation}
The gripper width therefore increases monotonically with the opening angle, ensuring a unique correspondence between them. At full opening, the maximum gripper width reaches $84.9\,\mathrm{mm}$, with a vertical displacement of $\delta=-14.9\,\mathrm{mm}$ relative to the closed position. This opening range closely matches the target parallel gripper's opening range.

Using these kinematic relationships, we analyze force transmission during a symmetric quasistatic grasp with the common base fixed and the links assumed rigid, neglecting friction, gravity, and tangential contact loads. Let $F_g$ denote the magnitude of the contact force acting on each side of the gripper, $\tau_h$ the balancing torque about each handle pivot, and $F_h$ the magnitude of the hand force applied at the corresponding grip point on the handle. The effective moment arm $\ell_h$ is the perpendicular distance from the handle pivot to the line of action of $F_h$, giving $\tau_h=F_h\ell_h$. When the hand force is perpendicular to the line connecting the pivot and the grip point, $\ell_h$ equals the distance between these two points, as shown in Fig.~\ref{fig:mechanism}(b).

For a virtual opening displacement $\mathrm{d}q$, the two handle torques contribute a combined work magnitude of $2\tau_h\,\mathrm{d}q$. Each gripper finger moves by $\mathrm{d}w_I/2$, so the contact forces contribute an opposing work magnitude of $F_g\,\mathrm{d}w_I$. Virtual work balance gives
\begin{equation}
    2\tau_h\,\mathrm{d}q=F_g\,\mathrm{d}w_I.
    \label{eq:virtual-work}
\end{equation}
Substituting~\eqref{eq:opening-jacobian} into~\eqref{eq:virtual-work} and using $\tau_h=F_h\ell_h$, we obtain
\begin{align}
    \tau_h
    &=\frac{F_g}{2}\frac{\mathrm{d}w_I}{\mathrm{d}q}
    =\rho F_g\sin(\varphi_0-q),
    \label{eq:handle-torque}\\
    \frac{F_h}{F_g}
    &=\frac{\rho\sin(\varphi_0-q)}{\ell_h}.
    \label{eq:force-ratio}
\end{align}
For $\ell_h=98\,\mathrm{mm}$, the predicted force ratios $F_h/F_g$ are $0.646$ at closure and $0.494$ at full opening. Measurements near these positions yielded $(F_g,F_h)=(3.9,2.6)\,\mathrm{N}$ and $(4.7,2.4)\,\mathrm{N}$, giving ratios of $0.667$ and $0.511$, respectively. The slightly higher measured ratios are consistent with joint friction, which the analysis neglects. Across the full opening range, the ratio varies only between $0.49$ and $0.65$, so the force perceived at the handles remains approximately proportional to the gripper contact force.

\subsection{Demonstration Collection System}
\label{sec:collection-system}

\subsubsection{Sensor Configuration and Calibration}
\label{sec:interface-calibration}

\begin{wrapfigure}{r}{0.49\textwidth}
    \centering
    \captionsetup{skip=5pt}
    \includegraphics[width=\linewidth]{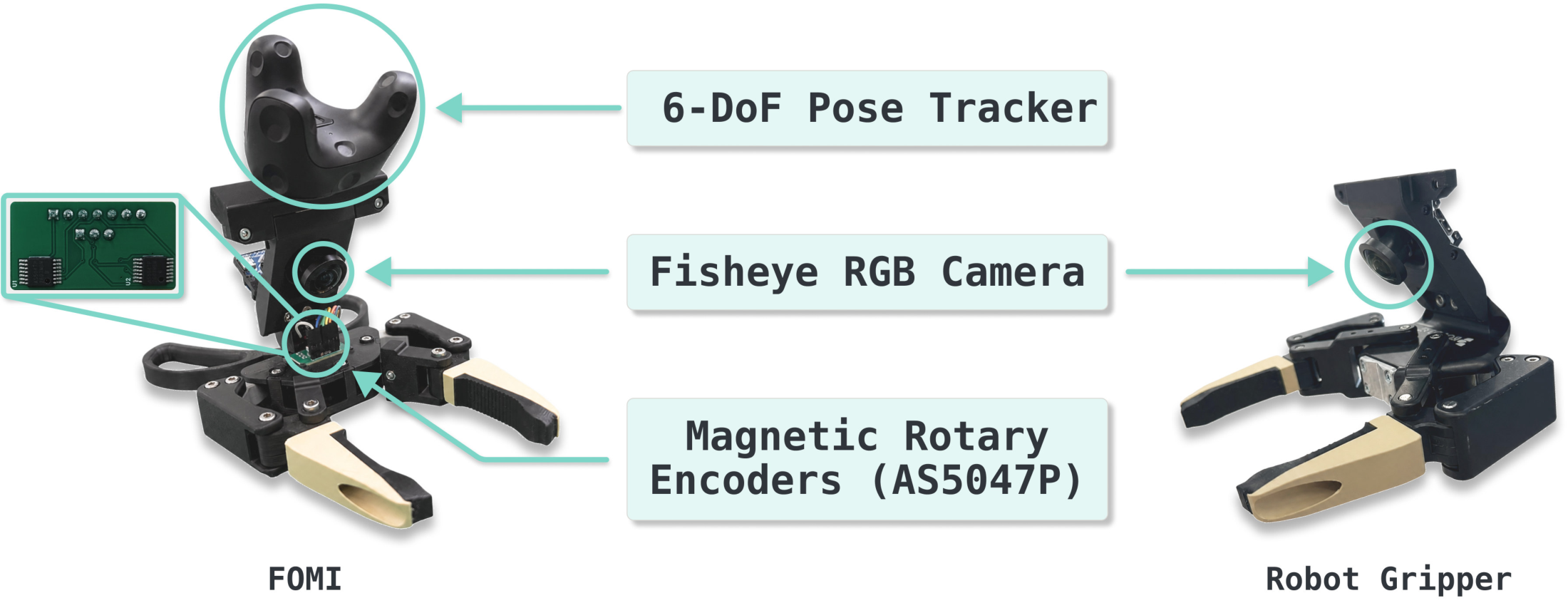}
    \caption{Left: sensor layout of FOMI. Right: the target robot gripper (Robotiq 2F-85).}
    \label{fig:sensors}
\end{wrapfigure}

As shown in Fig.~\ref{fig:sensors}, FOMI is configured with a fisheye RGB camera, a 6-DoF pose tracker, and two 14-bit magnetic rotary encoders (AS5047P). An identical fisheye camera is mounted on each robot wrist for teleoperation and policy deployment. Each encoder senses a permanent magnet embedded in the corresponding sector gear. An Arduino Nano board reads both encoders and outputs their wrap-corrected difference $x$, which is proportional to the opening coordinate $q$ in~\eqref{eq:width-kinematics} and serves as the gripper width signal. Encoder readings from both branches are used to reduce calibration errors associated with gear backlash.

For gripper-command calibration, we map the encoder signal $x$ to the target gripper's 8-bit position code $y\in[0,255]$, where $0$ and $255$ correspond to fully open and fully closed, respectively. A 3D-printed fixture mechanically couples FOMI to the robot gripper with their opening axes aligned. We command the robot gripper to uniformly spaced positions over the usable range in both opening and closing directions. After the motion settles at each position, we record $(x_n,y_n,\sigma_n)$, where $x_n$ is the encoder signal, $y_n$ is the robot gripper's reported position code, and $\sigma_n$ denotes the opening or closing direction.

Equations~\eqref{eq:width-kinematics} and~\eqref{eq:opening-jacobian} describe a nonlinear, monotonic relationship between gripper width and opening angle. Encoder readings also exhibit direction-dependent hysteresis. To account for nonlinearity and hysteresis, we fit
\begin{equation}
    f(x,\sigma)=p_k(x)+c_\sigma,
    \qquad x\in[\kappa_k,\kappa_{k+1}),
    \label{eq:width-model}
\end{equation}
where $p_k$ is a quadratic polynomial shared by both directions and $c_\sigma$ is a direction-dependent offset. The knots $\kappa_k$ are placed at the median encoder readings corresponding to uniformly spaced command levels. We estimate the polynomial coefficients and direction-dependent offsets by minimizing the least-squares objective $\sum_n [f(x_n,\sigma_n)-y_n]^2$, subject to $p_k(\kappa_{k+1})=p_{k+1}(\kappa_{k+1})$ to ensure continuity between adjacent segments. At run time, we determine the direction $\sigma_t$ from changes in the encoder signal over a sliding window of 5 samples to suppress direction changes caused by small fluctuations. The calibrated output $\hat{y}_t=f(x_t,\sigma_t)$ is then issued as the robot gripper command. Each interface is calibrated separately with its corresponding robot gripper.

For tracker-to-end-effector calibration, we define the end-effector frame $E$ at the tool center point (TCP). We temporarily transfer the tracker and its mount from FOMI to the camera bracket on the robot gripper. This mounting arrangement maintains the same tracker position and orientation relative to the end effector as on FOMI. We move the robot through a range of translations and rotations while recording synchronized end-effector and tracker poses. The end-effector pose in the robot base frame ${}^{B}\mathbf{T}_{E,i}$ is computed from robot forward kinematics and the known tool offset defining frame $E$. Together with the tracker pose ${}^{W}\mathbf{T}_{R,i}$, it satisfies
\begin{equation}
    {}^{B}\mathbf{T}_{E,i}=Y\,{}^{W}\mathbf{T}_{R,i}\,X,
    \label{eq:tracker-tcp-calibration}
\end{equation}
where $X={}^{R}\mathbf{T}_{E}$ maps the end-effector frame to the tracker frame and $Y={}^{B}\mathbf{T}_{W}$ maps the Lighthouse frame to the robot base frame. We estimate both transforms using robust nonlinear least squares; $Y$ is required only for this joint estimation and is not used thereafter. Calibration is accepted when the mean end-effector position and orientation reconstruction errors on held-out data are below $3\,\mathrm{mm}$ and $0.5^\circ$, respectively. After calibration, we reattach the tracker and its mount to FOMI and use the calibrated transform $X$ to compute the end-effector pose ${}^{W}\mathbf{T}_{E,t}={}^{W}\mathbf{T}_{R,t}X$ in both collection modes.

\subsubsection{Robot-Free Collection}
\label{sec:robot-free-collection}

During robot-free collection, the human operator uses the handheld FOMI interfaces, one in each hand, to manipulate objects, relying on intuitive force feedback to perceive contact and adjust actions in real time.
For each interface, each demonstration records images from its fisheye camera, the calibrated end-effector pose ${}^{W}\mathbf{T}_{E,t}$, and the calibrated gripper command $\hat{y}_t$.

\subsubsection{Teleoperation}
\label{sec:teleoperation}

In teleoperation, FOMI controls the robot online through the calibrated pose and gripper signals. At the first frame of each episode ($t=0$), we record the interface pose ${}^{W}\mathbf{T}_{E,0}$ and the measured robot end-effector pose ${}^{B}\mathbf{T}_{E,0}$, and compute the target pose at each subsequent time step as
\begin{equation}
    {}^{B}\mathbf{T}^{*}_{E,t}
    =
    {}^{B}\mathbf{T}_{E,0}
    \left({}^{W}\mathbf{T}_{E,0}\right)^{-1}
    {}^{W}\mathbf{T}_{E,t}.
    \label{eq:teleoperation-mapping}
\end{equation}
Here, $\left({}^{W}\mathbf{T}_{E,0}\right)^{-1}{}^{W}\mathbf{T}_{E,t}$ represents the interface's relative motion expressed in its initial end-effector frame. For bimanual teleoperation, each arm is controlled by a separate interface, with~\eqref{eq:teleoperation-mapping} applied independently using the corresponding initial interface and robot poses. The calibrated output $\hat{y}_t$ is used directly as the gripper width command. The operator relies on visual feedback, while FOMI's mechanical stops provide a physical indication of gripper width command limits without conveying object interaction forces. Each demonstration records images from the robot wrist cameras, the measured robot end-effector pose ${}^{B}\mathbf{T}_{E,t}$, and the calibrated gripper command $\hat{y}_t$.

\subsubsection{Data Synchronization}
\label{sec:synchronization}

The fisheye camera operates at $100\,\mathrm{Hz}$, the pose tracker at $120\,\mathrm{Hz}$, and the magnetic encoders are sampled at $200\,\mathrm{Hz}$. The acquisition workstation assigns timestamps to all data streams using a common clock. Before synchronization, timestamp gaps exceeding $20\,\mathrm{ms}$ between consecutive images are flagged as frame loss. These streams are aligned to an $80\,\mathrm{Hz}$ reference timeline by selecting the nearest sample in time from each stream. We record and visualize timestamp differences between data streams to check temporal alignment.
We exclude episodes with dropped frames flagged by the timestamp criterion or with tracking failures detected through visual inspection of the synchronized trajectories.

\subsection{Policy Learning and Deployment}
\label{sec:policy-learning}

For policy learning, we fine-tune the $\pi_{0.5}$ vision-language-action (VLA) model~\cite{intelligence2025pi} separately on robot-free and teleoperated demonstrations. The synchronized demonstration data are downsampled from $80$ to $20\,\mathrm{Hz}$ for policy learning. For training in both modes, poses in the observation history and future action targets are expressed relative to the left end-effector frame at the current observation time. This reference frame is held fixed for each action chunk. Gripper commands are represented as calibrated absolute values rather than relative increments. The policy predicts a bimanual action chunk of $H=50$ steps:
\begin{equation}
    \hat{\mathbf{a}}_{t:t+H-1}
    \sim \pi_{\theta}
    \big(\cdot\mid\mathbf{I}_t,\ell,\mathbf{s}_t\big),
    \label{eq:vla-policy}
\end{equation}
where $\mathbf{I}_t$ contains the two camera views, $\ell$ is the task instruction, and $\mathbf{s}_t\in\mathbb{R}^{32}$ concatenates the two most recent bimanual states. Each per-arm state contains a position, a unit quaternion, and a calibrated gripper command. Each action $\mathbf{a}_k\in\mathbb{R}^{16}$ contains the target poses and gripper commands for both arms.

At deployment, predicted poses are transformed into robot base coordinates using the reference pose fixed for each action chunk. For subsequent RTC chunks, this reference is the first left-arm target in the retained overlap; coordinate transformations and chunk switching are detailed in Appendix~\ref{app:rtc-implementation}. Real-time chunking~\cite{black2026real} allows committed actions to execute during inference while generating subsequent actions conditioned on the committed prefix. State histories are updated using robot forward kinematics and recorded gripper commands.


\section{Experiments}
\label{sec:experiments}

We first introduce the tasks and robot platforms (Sec.~\ref{sec:experimental-setup}) and then address two questions: \textbf{Q1:} Does FOMI improve demonstration collection efficiency and completion rate compared with baseline interfaces in both collection modes (Sec.~\ref{sec:collection-efficiency})?
\textbf{Q2:} Can demonstrations collected with FOMI train policies that perform contact-rich manipulation with high success rates (Secs.~\ref{sec:policy-evaluation} and~\ref{sec:cross-embodiment})?
We additionally evaluate the gripper-command calibration accuracy in Sec.~\ref{sec:encoder-calibration-experiment}.

\subsection{Tasks and Robot Platforms}
\label{sec:experimental-setup}

We use four real-world contact-rich tasks to evaluate demonstration collection efficiency, completion rate, and policy performance:
\begin{itemize}
    \item \textbf{Pick Chip} (Franka): Pick and place two fragile chips through delicate handling without breaking them.
    \item \textbf{Insert Tube} (Franka): Reorient a tube within the gripper, transfer it between the two grippers, and insert it with precision.
    \item \textbf{Wash Dish} (Franka): Wash a dish through sustained wiping contact, then place it down.
    \item \textbf{Open Pouch} (Franka and Flexiv): Open a pouch's zipper while accommodating the pouch's deformability.
\end{itemize}
These tasks require operators to adjust gripper opening and end-effector motion in response to object interaction forces during demonstration collection.
Both platforms are dual-arm setups that use identical Robotiq 2F-85 robot grippers and wrist cameras.

\subsection{Demonstration Collection Performance}
\label{sec:collection-efficiency}

\begin{wrapfigure}{r}{0.49\textwidth}
    \centering
    \captionsetup{skip=5pt}
    \includegraphics[width=\linewidth]{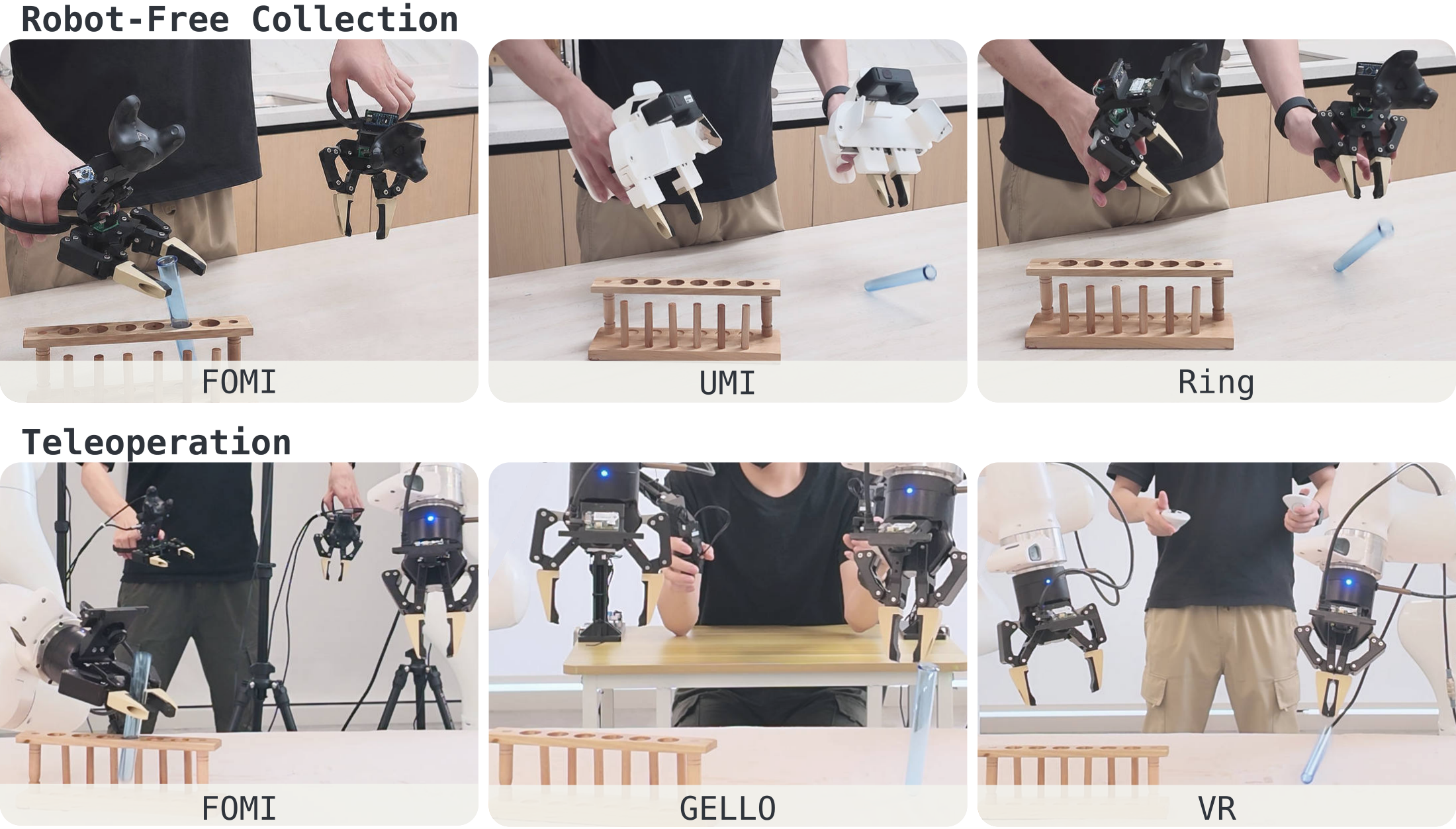}
    \caption{Demonstration collection interfaces and experimental setup used across all four tasks, illustrated using Insert Tube. Top: robot-free collection interfaces. Bottom: teleoperation interfaces.}
    \label{fig:collection-interfaces}
\end{wrapfigure}

We conducted an exploratory within-subject evaluation with three operators who had no prior robot operation experience operating robots in Fig.~\ref{fig:collection-interfaces}. Each operator used every interface on all four tasks, with interface and task orders randomized.
Sessions were repeated measurements from the same three operators.
For each interface and task combination, each operator completed three 10-minute collection sessions with 2-minute breaks between sessions.
An attempt was counted as a successful demonstration if all task stages were completed.
Collection efficiency is measured by the number of successful demonstrations per 10-minute session, averaged over the three sessions for each operator and then across operators.
For each interface, the completion rate is calculated as the total number of successful demonstrations divided by the total number of attempts across operators, sessions, and tasks.

\subsubsection{Robot-Free Collection Efficiency}

\begin{table}[!t]
\caption{Demonstration collection efficiency (successful demonstrations / 10 min). Entries report mean and sample standard deviation across three operators, with each operator's results averaged over three sessions.}
\label{tab:collection-efficiency}
\centering
\footnotesize
\setlength{\tabcolsep}{4pt}
\renewcommand{\arraystretch}{1.30}
\newcommand{\meanSD}[2]{$#1 \pm #2$}
\begin{tabular*}{\columnwidth}{@{\extracolsep{\fill}}l ccc ccc@{}}
\toprule
\multirow{2}{*}{\textbf{Task}} & \multicolumn{3}{c}{\textbf{Robot-Free} $\uparrow$} & \multicolumn{3}{c}{\textbf{Teleoperation} $\uparrow$} \\
\cmidrule(lr){2-4}\cmidrule(lr){5-7}
& UMI & Ring & FOMI & GELLO & VR & FOMI \\
\midrule
Pick Chip & \meanSD{16.6}{1.8} & \meanSD{10.7}{3.5} & \meanSD{\mathbf{22.0}}{1.5} & \meanSD{6.4}{0.8} & \meanSD{0.2}{0.4} & \meanSD{\mathbf{7.2}}{1.2} \\
Insert Tube & \meanSD{13.8}{1.0} & \meanSD{9.9}{3.0} & \meanSD{\mathbf{17.7}}{1.5} & \meanSD{4.9}{0.5} & \meanSD{1.6}{0.7} & \meanSD{\mathbf{7.4}}{1.3} \\
Wash Dish & \meanSD{11.0}{1.8} & \meanSD{10.9}{1.9} & \meanSD{\mathbf{13.3}}{2.4} & \meanSD{8.8}{0.4} & \meanSD{7.2}{0.4} & \meanSD{\mathbf{9.3}}{0.7} \\
Open Pouch & \meanSD{13.7}{3.4} & \meanSD{15.0}{3.0} & \meanSD{\mathbf{18.2}}{2.4} & \meanSD{8.1}{0.4} & \meanSD{4.6}{1.3} & \meanSD{\mathbf{9.7}}{1.7} \\
\midrule
\textbf{Mean} & $13.8$ & $11.6$ & $\mathbf{17.8}$ & $7.1$ & $3.4$ & $\mathbf{8.4}$ \\
\bottomrule
\end{tabular*}
\end{table}

We compare FOMI with UMI~\cite{chi2024umi} and Ring, our implementation of a finger-driven interface similar to FreeTacMan~\cite{wu2025freetacman} (Fig.~\ref{fig:collection-interfaces}, top).
As shown in Table~\ref{tab:collection-efficiency}, FOMI achieves the highest mean collection efficiency on all four tasks, improving the average efficiency by 29.0\% over UMI (17.8 vs.\ 13.8). We attribute this improvement to FOMI's combination of intuitive force feedback and handle support: UMI's trigger-driven rack-and-pinion mechanism transmits interaction forces through multiple intermediate stages, whereas Ring requires the actuating fingers to also support the interface.

\subsubsection{Teleoperation Efficiency}

We compare FOMI with GELLO~\cite{wu2024gello}, which provides joint-space control, and a VR interface using PICO controllers for end-effector pose and gripper width control, following OpenVR~\cite{george2025openvr} (Fig.~\ref{fig:collection-interfaces}, bottom).
As shown in Table~\ref{tab:collection-efficiency}, FOMI achieves the highest mean collection efficiency on all four tasks, improving the average efficiency by 18.3\% over GELLO (8.4 vs.\ 7.1).
We attribute this improvement to FOMI's gripper control: its opening directly corresponds to the robot gripper width, with mechanical stops indicating the command limits, whereas GELLO and VR control the gripper through handle opening angle and trigger travel, respectively.

\subsubsection{Completion Rate}

\begin{wrapfigure}{r}{0.49\textwidth}
    \centering
    \captionsetup{skip=5pt}
    \includegraphics[width=\linewidth]{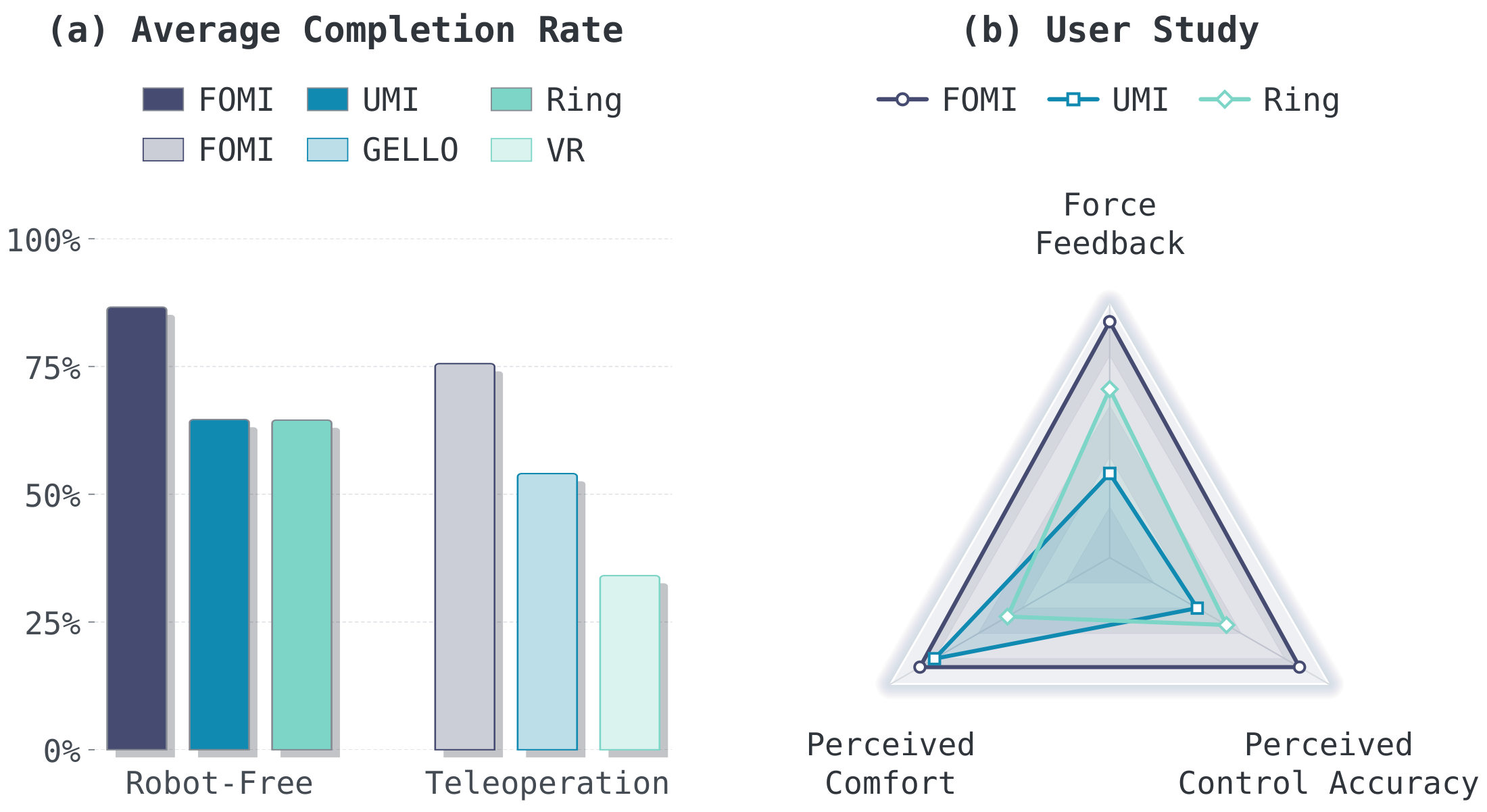}
    \caption{(a)~Completion rate, computed as total successful attempts divided by total attempts over tasks, sessions, and operators. (b)~Exploratory user study for the interfaces in robot-free collection: mean ratings from three operators across three dimensions, each rated on a 0--5 scale.}
    \label{fig:collection-efficiency}
\end{wrapfigure}

As shown in Fig.~\ref{fig:collection-efficiency}(a), FOMI achieves the highest completion rate among the evaluated interfaces in both collection modes.
In robot-free collection, its completion rate is 86.6\%, compared with 64.6\% for UMI and 64.5\% for Ring. In teleoperation, FOMI achieves 75.6\%, compared with 54.0\% for GELLO and 34.1\% for VR.
These results suggest that FOMI's efficiency gains partly stem from fewer failed attempts, as intuitive force feedback in robot-free collection and physical indications of gripper width limits provided by mechanical stops in teleoperation support real-time action adjustment.

\subsubsection{Exploratory User Study in Robot-Free Collection}
\label{sec:user-study}

After completing the robot-free collection experiments, each operator rated FOMI, UMI, and Ring on perceived comfort, control accuracy, and force feedback using a 0--5 scale (Fig.~\ref{fig:collection-efficiency}(b)).
All three operators rated FOMI higher than both baselines for control accuracy and force feedback, with mean ratings of 4.33 and 4.67 for FOMI, compared with at most 2.67 and 3.33 for the baselines. FOMI and UMI received similar comfort ratings (4.33 and 4.00), whereas Ring received 2.33, consistent with its lack of handle support. These ratings provide preliminary evidence that operators perceived FOMI's intuitive force feedback and control more favorably.

\WFclear

\begin{figure}[H]
    \centering
    \includegraphics[width=1.0\textwidth]{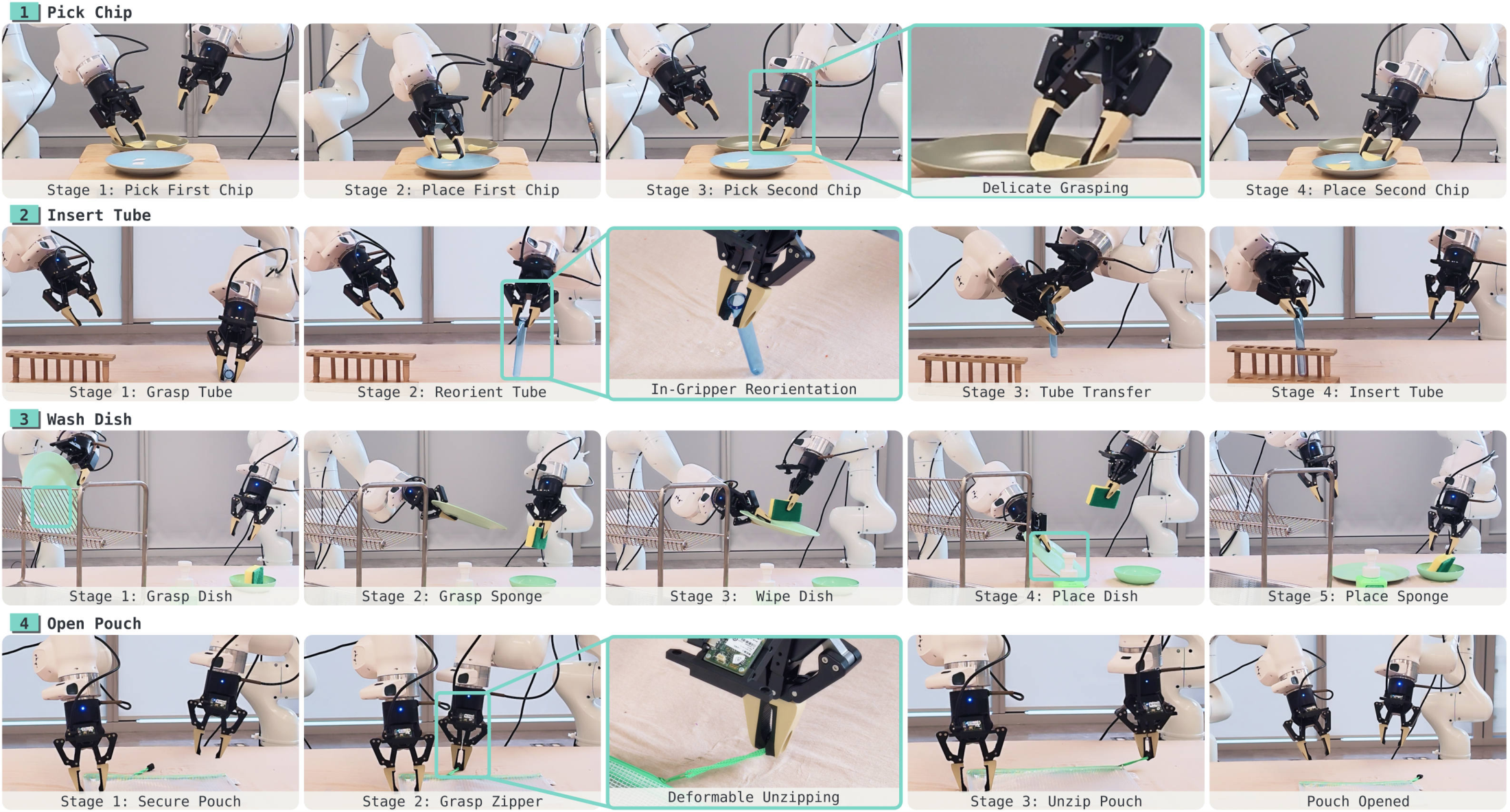}
    \caption{Representative rollouts of policies trained on robot-free demonstrations across four tasks. Labeled stages are used to compute progress rate. Insets highlight details of contact-rich manipulation.}
    \label{fig:policy}
\end{figure}

\subsection{Policy Performance with FOMI}
\label{sec:policy-evaluation}

\begin{wrapfigure}{r}{0.49\textwidth}
    \centering
    \captionsetup{skip=5pt}
    \includegraphics[width=\linewidth]{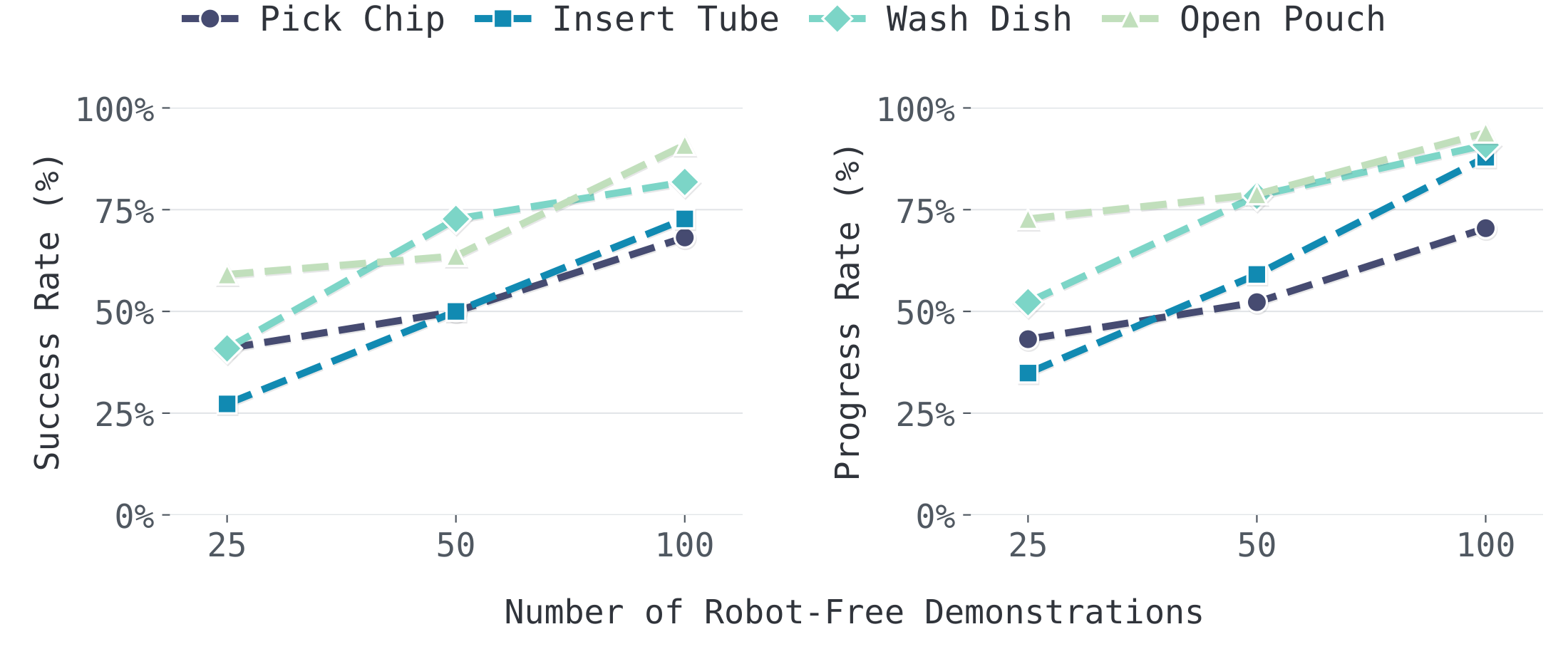}
    \caption{Policy performance with increasing numbers of robot-free demonstrations per task: 25, 50, and 100. Left: success rate. Right: progress rate.}
    \label{fig:robot-free-policy-scaling}
\end{wrapfigure}

For each task, we fine-tune a policy for 10,000 steps with a batch size of 64.
Each policy predicts an action chunk of $H=50$ steps, executed at $20\,\mathrm{Hz}$.
During inference, the first three actions remain fixed and continue to execute as the committed prefix in real-time chunking, accommodating an inference latency of approximately $120\,\mathrm{ms}$.
We evaluate each condition (collection mode and dataset size) over 22 rollouts per task, with matched initial object configurations across policies.
Success rate is the percentage of rollouts that complete all task stages.
For example, any chip breakage in Pick Chip counts as a failure.
Progress rate averages each rollout's stage scores over the stages in Fig.~\ref{fig:policy} and then across rollouts. Stages receive 1 point for first-attempt completion, 0.5 after a retry, and 0 otherwise.

\WFclear
\subsubsection{Robot-Free Demonstrations}

We fine-tune the $\pi_{0.5}$ policy separately using 25, 50, and 100 robot-free demonstrations per task, with each smaller dataset randomly sampled from the next larger dataset.
As shown in Fig.~\ref{fig:robot-free-policy-scaling}, increasing the number of demonstrations raises the average success rate from 42.0\% (25) to 59.1\% (50) and 78.4\% (100), and the average progress rate from 50.8\% to 67.1\% and 85.8\%.
At the average collection efficiency, collecting 100 demonstrations requires approximately one hour of active collection. With 100 demonstrations per task, policies achieve high success rates on the evaluated contact-rich tasks, and the consistent gains suggest further improvement with more robot-free data.

\WFclear
\Needspace{22\baselineskip}
\subsubsection{Teleoperated Demonstrations}

\noindent
\begin{minipage}[t]{0.47\textwidth}
\vspace{0pt}
As shown in Fig.~\ref{fig:teleoperation-policy}, the $\pi_{0.5}$ policy fine-tuned on 25 teleoperated demonstrations per task achieves average success and progress rates of 70.5\% and 83.0\%.
At the same dataset size, the success rate exceeds the 42.0\% obtained with robot-free demonstrations, possibly because teleoperated data are collected on the target robot without visual or kinematic gaps. Robot-free collection, however, is more than twice as efficient (17.8 vs.\ 8.4 demonstrations per 10 minutes) and requires no robot hardware.
\end{minipage}\hfill
\begin{minipage}[t]{0.49\textwidth}
    \vspace{0pt}
    \centering
    \captionsetup{skip=5pt}
    \includegraphics[width=\linewidth]{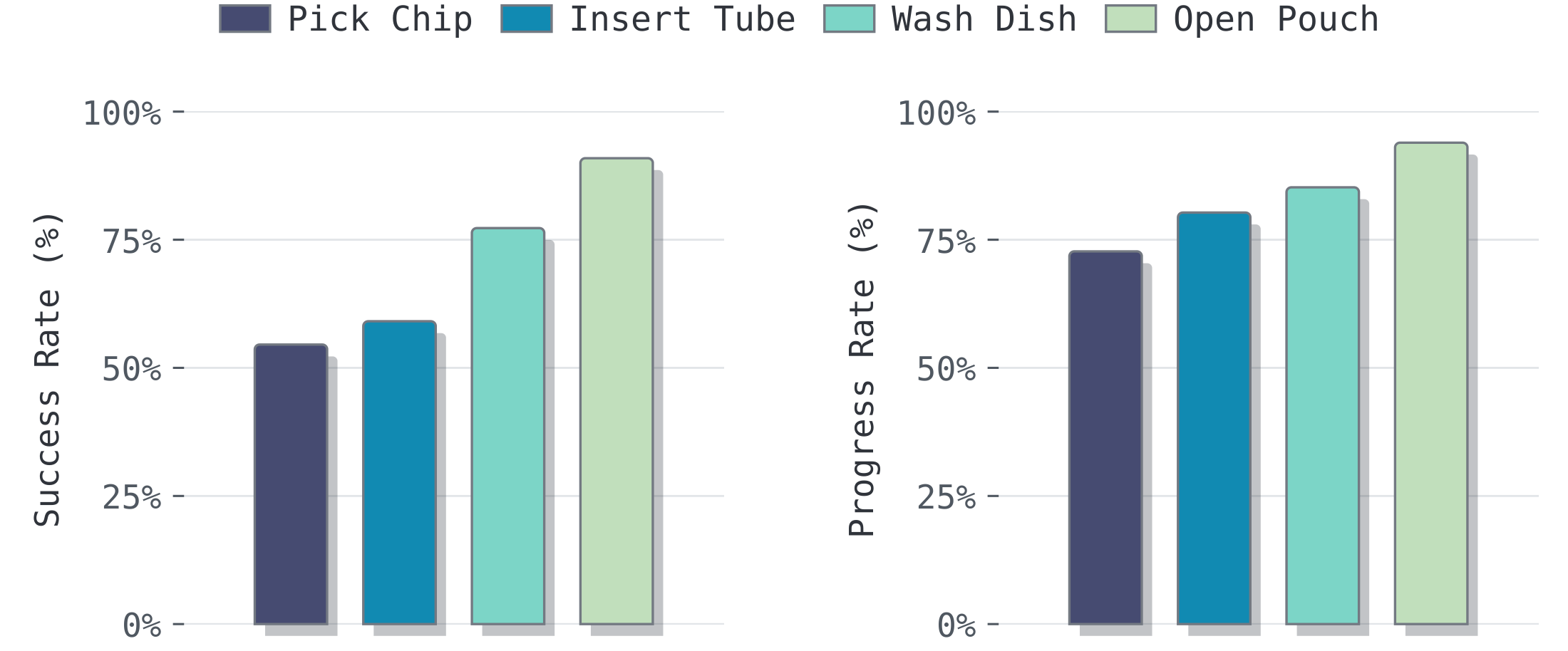}
    \captionof{figure}{Policy performance with 25 teleoperated demonstrations per task. Left: success rate. Right: progress rate.}
    \label{fig:teleoperation-policy}
\end{minipage}
\par

\subsection{Deployment Across Robot Embodiments}
\label{sec:cross-embodiment}

\begin{figure}[!htbp]
    \centering
    \includegraphics[width=0.78\columnwidth]{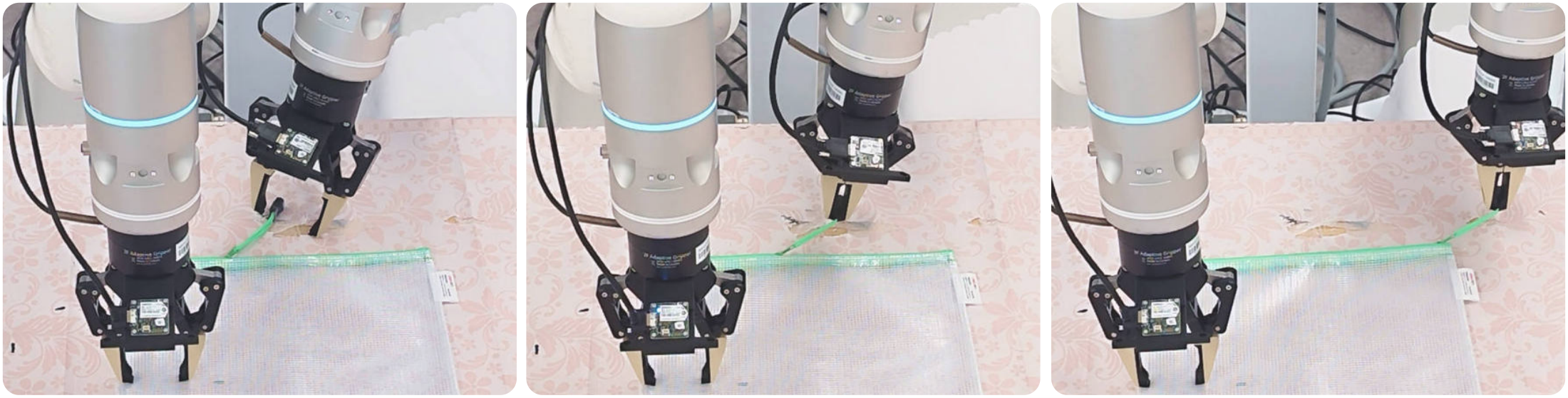}
    \caption{Policy performance on Open Pouch after deploying the policy fine-tuned on 25 robot-free demonstrations to the Flexiv platform.}
    \label{fig:multi-embodiment-deployment}
\end{figure}

We further evaluate whether robot-free demonstrations collected with FOMI transfer across robot platforms.
We deploy the Open Pouch policy trained on 25 robot-free demonstrations directly on a dual-arm Flexiv platform, achieving success and progress rates of 68.2\% and 80.3\%, respectively (Fig.~\ref{fig:multi-embodiment-deployment}).
These results support the reuse of FOMI's robot-free demonstrations across the evaluated robot platforms, which share the same gripper and wrist camera. The demonstrations encode relative end-effector poses and gripper commands without relying on arm-specific joint configurations.

\subsection{Gripper-Command Calibration Accuracy}
\label{sec:encoder-calibration-experiment}

\begin{wraptable}{r}{0.49\textwidth}
    \caption{Gripper-Command Calibration RMSE (Command Range: 0--255)}
    \label{tab:encoder-calibration}
    \centering
    \footnotesize
    \begin{tabularx}{\linewidth}{@{}>{\raggedright\arraybackslash}Xcc@{}}
        \toprule
        \textbf{Model} & \textbf{Left} $\downarrow$ & \textbf{Right} $\downarrow$ \\
        \midrule
        Linear regression & 3.36 & 4.47 \\
        MLP & 0.99 & 0.94 \\
        Piecewise quadratic, without offset & 2.48 & 3.49 \\
        Piecewise quadratic, with offset (ours)
        & \textbf{0.80} & \textbf{0.88} \\
        \bottomrule
    \end{tabularx}
\end{wraptable}

We evaluate the gripper-command mapping described in Sec.~\ref{sec:interface-calibration} on the left and right FOMI interfaces.
For each interface, we separately collect 2,000 fitting samples and 400 test samples.
We compare four models: linear regression $\hat{y}=ax+b$; a multilayer perceptron (MLP); the piecewise quadratic mapping with $c_\sigma=0$; and the piecewise quadratic mapping $f(x,\sigma)=p_k(x)+c_\sigma$ with a direction-dependent offset.
The MLP has two hidden layers with 16 and 8 units and takes the normalized encoder signal, inferred opening or closing direction, and a 15-bin one-hot position encoding as inputs.
Its results are averaged over three random seeds.
Direction is inferred online using the procedure described in Sec.~\ref{sec:interface-calibration}.
All hyperparameters are selected using only the fitting data.

As shown in Table~\ref{tab:encoder-calibration}, the piecewise quadratic mapping with a direction-dependent offset achieves the lowest RMSE on both interfaces: 0.80 command units ($\approx 0.3\,\mathrm{mm}$) on the left and 0.88 ($\approx 0.3\,\mathrm{mm}$) on the right, both below 0.35\% of the full command range.
Removing the offset increases the errors to 2.48 and 3.49, respectively, demonstrating its contribution to the calibration accuracy.
Although the MLP achieves comparable accuracy, the piecewise quadratic mapping is more interpretable and requires fewer parameters.


\section{Conclusion}

We presented FOMI, a scissor-shaped manipulation interface, and a demonstration collection system built around it. The interface provides intuitive mechanical force feedback during robot-free collection, while mechanical stops indicate gripper width command limits in teleoperation. FOMI achieves higher collection efficiency and completion rates than the evaluated baselines under the tested conditions. Demonstrations collected in both modes support policy learning for the evaluated contact-rich tasks without tactile or force inputs. However, interaction forces are not recorded for policy learning, and the limited sample size of the user study limits generalization across operators. Future work will expand the user study and refine the mechanical design. We will also explore additional sensory inputs for policy learning from demonstrations collected with intuitive force feedback.


\clearpage
\bibliography{references}
\bibliographystyle{template}

\clearpage
\appendix
\section*{Appendix}

\section{Force Transmission Measurement Setup}
\label{app:force-measurement}

For the force measurements, the common base is fixed while the links and handles remain free to move. Two tension force gauges are used: one measures the force $F_h$ at the designated grip point on one handle, and the other is coupled to the gripper contact location through a cord and fixture to measure the applied force $F_g$. Both force readings are recorded together with the corresponding opening angle $q$ measured by the encoders.

\section{Image Views During Demonstration Collection}
\label{app:image-views}

\begin{figure}[!htbp]
    \centering
    \includegraphics[width=1.0\linewidth]{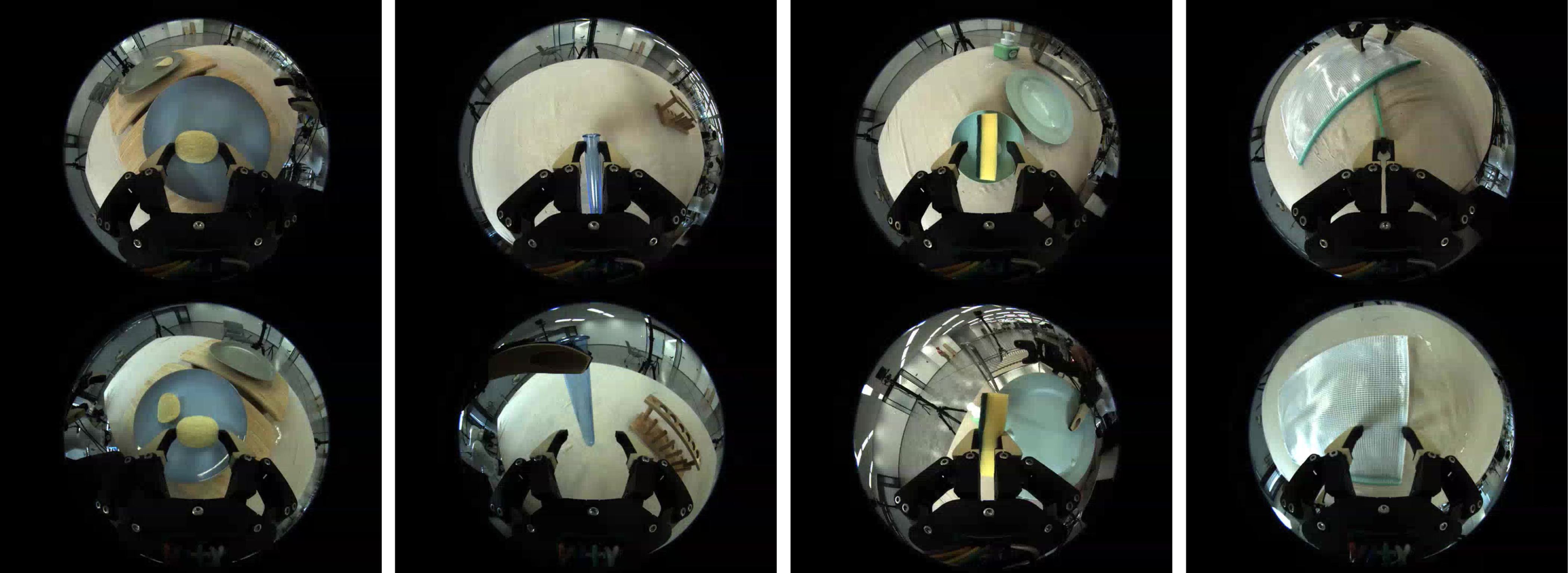}
    \caption{Representative camera views recorded during demonstration collection. From left to right: Pick Chip, Insert Tube, Wash Dish, and Open Pouch.}
    \label{fig:collection-image-views}
\end{figure}

Figure~\ref{fig:collection-image-views} shows representative fisheye camera images recorded during demonstration collection for the four contact-rich tasks. These images capture the grippers and object interactions and serve as visual observations for policy learning. FOMI closely matches the robot gripper in geometry and visual appearance.

\section{Demonstration Collection Protocol and Baseline Settings}
\label{app:collection-protocol}

Each collection session includes 10 minutes of active demonstration collection, excluding time spent on object resetting and failure recovery. An attempt is considered unsuccessful if the task setup is altered such that further retries are no longer possible, such as when a chip breaks, or if more than two retries are required.

For robot-free collection, UMI uses a GoPro camera, while Ring uses the same camera and pose tracker as FOMI. All three interfaces have a maximum gripper opening of approximately $84.9\,\mathrm{mm}$. For teleoperation, the VR interface maps trigger travel to robot gripper width commands, while GELLO maps the handle opening angle to these commands. Table~\ref{tab:collection-counts} reports the successful and failed attempts underlying the completion rates in Fig.~\ref{fig:collection-efficiency}(a). Completion rates are computed from pooled counts as $100S/(S+F)$.

\begin{table}[!htbp]
\caption{Demonstration collection attempt counts. Each task row pools three sessions from each of three operators. $S$: successful attempts; $F$: failed attempts. Bold marks the highest success and lowest failure counts within each mode.}
\label{tab:collection-counts}
\centering
\footnotesize
\setlength{\tabcolsep}{3pt}
\renewcommand{\arraystretch}{1.20}
\begin{tabular*}{0.98\linewidth}{@{\extracolsep{\fill}}l *{12}{c}@{}}
\toprule
\multirow{3}{*}{\textbf{Task}} & \multicolumn{6}{c}{\textbf{Robot-Free}} & \multicolumn{6}{c}{\textbf{Teleoperation}} \\
\cmidrule(lr){2-7}\cmidrule(lr){8-13}
& \multicolumn{2}{c}{UMI} & \multicolumn{2}{c}{Ring} & \multicolumn{2}{c}{FOMI} & \multicolumn{2}{c}{GELLO} & \multicolumn{2}{c}{VR} & \multicolumn{2}{c}{FOMI} \\
\cmidrule(lr){2-3}\cmidrule(lr){4-5}\cmidrule(lr){6-7}\cmidrule(lr){8-9}\cmidrule(lr){10-11}\cmidrule(lr){12-13}
& $S\!\uparrow$ & $F\!\downarrow$ & $S\!\uparrow$ & $F\!\downarrow$ & $\boldsymbol{S}\!\uparrow$ & $\boldsymbol{F}\!\downarrow$ & $S\!\uparrow$ & $F\!\downarrow$ & $S\!\uparrow$ & $F\!\downarrow$ & $\boldsymbol{S}\!\uparrow$ & $\boldsymbol{F}\!\downarrow$ \\
\midrule
Pick Chip & 149 & 71 & 96 & 66 & \textbf{198} & \textbf{28} & 58 & 70 & 2 & 92 & \textbf{65} & \textbf{32} \\
Insert Tube & 124 & 61 & 89 & 57 & \textbf{159} & \textbf{18} & 44 & 83 & 14 & 85 & \textbf{67} & \textbf{36} \\
Wash Dish & 99 & 51 & 98 & 39 & \textbf{120} & \textbf{24} & 79 & 28 & 65 & 13 & \textbf{84} & \textbf{11} \\
Open Pouch & 123 & 88 & 135 & 68 & \textbf{164} & \textbf{29} & 73 & 35 & 41 & 46 & \textbf{87} & \textbf{19} \\
\midrule
\textbf{Total} & 495 & 271 & 418 & 230 & \textbf{641} & \textbf{99} & 254 & 216 & 122 & 236 & \textbf{303} & \textbf{98} \\
\bottomrule
\end{tabular*}
\end{table}
\FloatBarrier

\section{Policy Training and Inference Details}
\label{app:policy-training}

\subsection{Training Details}
\label{app:training-details}

We initialize each policy from the $\pi_{0.5}$ base checkpoint~\cite{intelligence2025pi} and fine-tune it for 10,000 steps with a batch size of 64. We use AdamW with a peak learning rate of $1.5\times10^{-5}$, a 1,000-step warmup, and cosine decay. Each $720\times540$ image is center-cropped to $300\times300$ pixels and resized to $224\times224$ pixels for policy input. We use the final checkpoint at step 10,000 for evaluation, with 22 rollouts per task under each experimental condition.

\subsection{Real-Time Chunking}
\label{app:rtc-implementation}

At deployment, we use real-time chunking (RTC)~\cite{black2026real} and generate each action chunk using 10 flow-matching integration steps. We use an exponential weighting schedule for prefix guidance. Actions are executed at $20\,\mathrm{Hz}$, and a new inference call is triggered after every 20 executed actions. During inference, a committed prefix of 3 actions remains fixed and continues to execute, accommodating an inference latency of approximately $120\,\mathrm{ms}$.

The client stores each predicted trajectory as absolute TCP poses for both arms in the left robot base frame $B_L$, together with a copy expressed in each arm's own base frame for execution. After $s=20$ actions have been executed, the remaining $H-s=30$ actions form the overlap for the next inference call. Retaining absolute poses preserves the spatial targets when the reference frame changes.

Before inference, the overlap and the two-step history are re-expressed relative to the current left TCP pose. The server then uses the first left-arm target in the overlap as the model reference. Let $T_i^{B_L}$ denote an absolute target pose, and let $T_{\mathrm{current}}$ and $T_{\mathrm{anchor}}$ denote the current left TCP pose and the first left-arm overlap target, respectively, both expressed in $B_L$. The transformations are
\begin{align}
T_i^{\mathrm{current}} &= T_{\mathrm{current}}^{-1}T_i^{B_L},\\
T_i^{\mathrm{model}} &= T_{\mathrm{anchor}}^{-1}T_i^{B_L}.
\end{align}
Both arms use the same model reference, and gripper commands remain unchanged. Thus, for subsequent RTC chunks, the model reference is a retained target pose rather than necessarily the currently measured left TCP pose. The client retains this anchor for decoding the new chunk. Right-arm predictions are decoded relative to the first decoded left-arm target and then transformed into the right robot base frame.

While inference runs, the controller continues executing the retained overlap. When the new prediction arrives, the client uses the number of actions already consumed during inference, $d$, to align the replacement chunk with the execution progress. With zero-based indexing, it installs actions $d$ through $s-1$ from the new chunk, avoiding replay of elapsed actions. If inference exceeds the allowed switching deadline, the system falls back to synchronous replanning.

\end{document}